\documentclass[lettersize,journal]{IEEEtran}
\usepackage{amsmath,amsfonts}
\usepackage{algorithmic}
\usepackage{algorithm}
\usepackage{array}
\usepackage[caption=false,font=normalsize,labelfont=sf,textfont=sf]{subfig}
\usepackage{textcomp}
\usepackage{stfloats}
\usepackage{url}
\usepackage{verbatim}
\usepackage{graphicx}
\usepackage{cite}
\usepackage{xcolor}
\usepackage{booktabs}
\usepackage{multirow}
\usepackage{caption}  
\begin{document}

\title{Clustered Federated Learning for Generalizable FDIA Detection in Smart Grids with Heterogeneous Data}

\author{Yunfeng Li,~\IEEEmembership{Student Member,~IEEE}, Junhong Liu,~\IEEEmembership{Member,~IEEE}, Zhaohui Yang,~\IEEEmembership{Student Member,~IEEE}\\
Guofu Liao,~\IEEEmembership{Student Member,~IEEE}, Chuyun Zhang,~\IEEEmembership{Student Member,~IEEE}

\thanks{Yunfeng Li is with the Department of Computer Science, University of California, Santa Barbara, CA, USA  (e-mail: yunfengli@ucsb.edu).}
\thanks{Junhong Liu is with the Department of Electrical and Electronic Engineering, The University of Hong Kong, Hong Kong SAR, China (e-mail: jhliu@eee.hku.hk).}
\thanks{Zhaohui Yang is with the Department of Computer Science, University of California, Santa Barbara, CA, USA (e-mail: zhaohui@ucsb.edu).}
\thanks{Guofu Liao is with the Department of Electronics and
Information Engineering, Shenzhen University,
Shenzhen, China (e-mail: liaoguofu2022@email.szu.edu.cn).}
\thanks{Chuyun Zhang is with the Department of Electronics and
Information Engineering, Shenzhen University,
Shenzhen, China (e-mail: cyzhang@szu.edu.cn).}

}



\maketitle

\begin{abstract}
False Data Injection Attacks (FDIAs) pose severe security risks to smart grids by manipulating measurement data collected from spatially distributed devices such as SCADA systems and PMUs. These measurements typically exhibit Non-Independent and Identically Distributed (Non-IID) characteristics across different regions, which significantly challenges the generalization ability of detection models. Traditional centralized training approaches not only face privacy risks and data sharing constraints but also incur high transmission costs, limiting their scalability and deployment feasibility.
To address these issues, this paper proposes a privacy-preserving federated learning framework, termed Federated Cluster Average (FedClusAvg), designed to improve FDIA detection in Non-IID and resource-constrained environments. FedClusAvg incorporates cluster-based stratified sampling and hierarchical communication (client–subserver–server) to enhance model generalization and reduce communication overhead. By enabling localized training and weighted parameter aggregation, the algorithm achieves accurate model convergence without centralizing sensitive data.
Experimental results on benchmark smart grid datasets demonstrate that FedClusAvg not only improves detection accuracy under heterogeneous data distributions but also significantly reduces communication rounds and bandwidth consumption. This work provides an effective solution for secure and efficient FDIA detection in large-scale distributed power systems.

\end{abstract}


\section{Introduction}
\IEEEPARstart{A}{s} an important cyber-physical system (CPS), smart grid is highly vulnerable to cyber attacks \cite{li2022detection}. In modern smart grid infrastructures, measurement devices such as Supervisory Control and Data Acquisition (SCADA) systems and Phasor Measurement Units (PMUs) are extensively deployed across geographically distributed regions to monitor real-time operational status and collect critical system data. Due to variations in physical environments, network topologies, load profiles, and regional energy mixes, the acquired data is often highly heterogeneous, exhibiting pronounced spatio-temporal variability and non-independent and identically distributed (Non-IID) characteristics across regions, where even identically-named variables may follow distinct distributions~\cite{zhang2024federated,Shrestha2024Anomaly, liu2025byzantine}.

This statistical heterogeneity further extends to cyberattack behaviors, particularly in the context of false data injection attacks (FDIAs), where attack vectors demonstrate spatial patterns. For instance, adversaries in urban areas may compromise power flow data to induce incorrect economic dispatch decisions, while attackers in regions rich in renewable energy may manipulate forecast signals or distributed generator states~\cite{reda2022comprehensive,Sater2021Anomaly}. These non-uniform attack patterns pose significant challenges to traditional centralized machine learning-based detection frameworks, which assume homogenous training data.

Centralized FDIA detection models typically require aggregating sensitive grid data—such as substation measurements, load consumption profiles, and generation forecasts—into a centralized data center. However, such architectures face growing scrutiny due to the risks of privacy leakage \cite{liu2023privacy}, compliance violations (e.g., GDPR~\cite{regulation2016regulation}), and exposure to cyber-physical attacks or system-wide failures~\cite{Gooi2023Edge,Blika2024Federated}. Moreover, institutional and jurisdictional boundaries often inhibit direct data sharing among regional grid operators, reinforcing data silos and weakening the global generalizability of trained models.
The increasing deployment of edge computing platforms and the rising volume of sensor data also make centralized processing cost-prohibitive, suffering from latency bottlenecks and bandwidth limitations~\cite{Haghnegahdar2024SHAP}. These practical concerns underscore the urgent demand for privacy-preserving, bandwidth-efficient, and generalizable frameworks for FDIA detection under real-world heterogeneous conditions.

Federated Learning (FL) has emerged as a transformative paradigm that enables decentralized model training across distributed clients without exchanging raw data \cite{su2021secure}. Instead, clients periodically transmit encrypted or abstracted model updates-such as gradients or model weights-to a server for aggregation~\cite{yang2019federated,Hourixin2020PersonalPrivacy}. This architecture not only preserves local data privacy and aligns with regional regulations, but also supports scalability in geographically distributed energy systems~\cite{Latif2025SecFL}. 

To effectively detect FDIA in smart grids, researchers have developed several privacy-preserving FL approaches. These include implementations using the Paillier cryptosystem and Local Differential Privacy (LDP) techniques \cite{wen2021feddetect}. Additionally, decentralized FL frameworks have been specifically designed to identify false data injection attacks targeting solar PV dc/dc and dc/ac converters \cite{zhao2021federated}, as well as to detect fraudulent behavior among advanced metering infrastructure consumers who compromise their meters \cite{bondok2023novel}.
To enhance the robustness of FL-based machine learning models against adversarial attacks during both training and inference phases, researchers have explored novel FL-based false data detection methods incorporating Explainable Artificial Intelligence \cite{elgarhy2025investigation}. Furthermore, to address challenges posed by unknown system parameters and limited decentralized datasets with strategic data owners, incentive-based edge FL mechanisms have been developed \cite{lin2022incentive}.
Several advanced frameworks have emerged to address specific smart grid security challenges. These include communication-efficient and privacy-preserving FL schemes for secure energy data sharing through edge-cloud collaboration \cite{su2021secure}, and collaborative learning frameworks utilizing vertical FL for detecting false data injection attacks \cite{kesici2024detection}. To protect against inference attacks from potentially dishonest aggregators who might extract information about clients' training data from model parameters, efficient cross-silo FL schemes with enhanced privacy preservation have been developed \cite{tran2023efficient}. Moreover, researchers have developed FL schemes that exploits local correlations between connected power buses through graph neural networks while capturing temporal patterns using long short-term memory layers \cite{kecceci2025federated}. Nevertheless, FL suffers from inherent performance degradation under highly non-IID settings~\cite{Quan2025CPS}. To address this, Clustered Federated Learning (CFL) strategies have been proposed, which group clients based on statistical similarity or attack vector profiles, and train models within each cluster separately~\cite{Shabbir2024Lightweight,Ruan2023DeepLearning,Chifu2024FedWOA}. This clustering not only improves local model relevance but also enhances robustness to adversarial manipulations.

Although FL has demonstrated effectiveness in smart grid anomaly detection, several technical challenges persist~\cite{FederatedLearning_Grid_2022}:
\begin{itemize}
  \item \textbf{Non-IID Data:} Heterogeneous data distributions across clients can lead to model divergence and hinder global convergence.
  \item \textbf{Communication Overhead:} Frequent communication between clients and the central server imposes latency and increases bandwidth usage.
  \item \textbf{Resource Constraints:} Many edge devices have limited computation and memory capacities, making local training of large-scale models challenging.
\end{itemize}

To address these limitations, this paper proposes a novel algorithm termed Federated Clustered Averaging (FedClusAvg), specifically designed for FDIA detection under Non-IID conditions and stringent privacy constraints. The proposed framework adopts a hierarchical communication architecture (client-subserver-server) and employs cluster-based model aggregation to enhance training efficiency, reduce communication costs, and improve model generalization in distributed energy environments.

\begin{figure}[t] 
	\centering
	\includegraphics[width=0.7\linewidth]{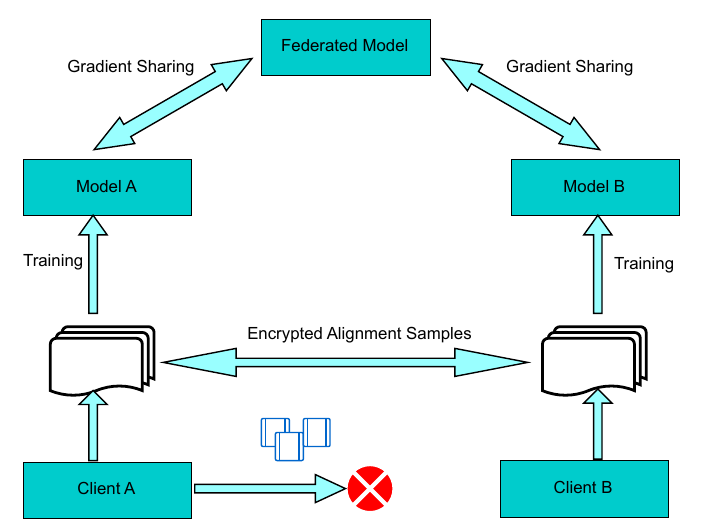}
	\caption{Federated Learning Framework}
	\label{fig:11}
\end{figure}

The algorithm in this paper can be applied in the following form: the private data of each client cannot be exchanged, the samples of each client are encrypted and aligned, a model is trained locally and then the model parameters are updated through gradient sharing to achieve the purpose of the federated training mode. Its form is shown in Fig \ref{fig:11}.
The primary innovations of FedClusAvg are summarized as follows:
\begin{enumerate}
  \item \textbf{Clustered Parameter Aggregation:} Clients are grouped based on the similarity of data distributions. Within each cluster, local models are aggregated using deviation-weighted strategies to mitigate client drift and improve convergence under Non-IID conditions.
  \item \textbf{Hierarchical Communication Structure:} Intermediate sub-servers first perform cluster-level aggregation before forwarding results to the central server. This significantly reduces communication rounds and improves scalability in large-scale systems.
  \item \textbf{Privacy Preservation:} Only model parameters are shared across communication links. The framework is extendable to support differential privacy and secure multi-party computation for enhanced data protection.
  \item \textbf{Resource Efficiency and Scalability:} The design supports multi-epoch local training, minimizes communication overhead, and facilitates deployment on resource-constrained edge devices such as intelligent substations or field gateways.
\end{enumerate}

By integrating federated learning with clustered model aggregation, FedClusAvg enhances model robustness and accuracy in the presence of highly heterogeneous and geographically biased data, while simultaneously reducing training time and communication latency. These capabilities make it particularly well-suited for FDIA detection tasks that demand real-time decision-making and stringent privacy safeguards.


\section{Federated Clustered Averaging for FDIA Detection} \label{Intro1}

Most successful applications of federated learning (FL) rely on variants of stochastic gradient descent (SGD). A foundational optimization algorithm in FL is the Federated Averaging (FedAvg) algorithm, first introduced by McMahan et al.~\cite{mcmahan2017communication}. FedAvg enables clients to perform multiple local SGD updates in parallel before synchronizing with a central server, thereby substantially reducing the number of communication rounds required for global model convergence.

Although FedAvg partially alleviates the impact of data heterogeneity (Non-IID distributions), its performance often degrades significantly when client data distributions differ considerably. In such scenarios, direct aggregation of local updates—without accounting for divergence between client and global models—can lead to reduced accuracy and unstable convergence.

To address this limitation, we interpret the deviation of each client’s model parameters from the global average as a proxy for the reliability of its contribution. Clients exhibiting large deviations are more likely to introduce bias and are therefore assigned lower weights during aggregation, improving model robustness under heterogeneous conditions.

However, empirical evidence suggests that client-level data heterogeneity does not always align with observed parameter deviations. To resolve this, we propose a sample-level clustering strategy that partitions clients with high internal data variability into smaller virtual sub-clients based on feature similarity. These sub-clients undergo hierarchical stratified sampling to construct local training sets with reduced intra-client divergence.

This resampling mechanism effectively mitigates the internal distributional variance within each client, thereby stabilizing the training dynamics under Non-IID conditions. Following local training, we apply a weighted aggregation scheme during model synchronization. Aggregation weights are determined based on training quality and parameter consistency, enabling more accurate and adaptive global model updates.

Collectively, the combination of sample-level clustering and adaptive weighted aggregation significantly enhances convergence speed and model performance in federated environments with highly heterogeneous data.

\subsection{Algorithmic Framework Design}

Building upon the classical Federated Averaging (FedAvg) algorithm, we propose an enhanced approach termed Federated Clustered Averaging (FedClusAvg). While retaining the conventional two-tier federated learning architecture, comprising clients and a central server, FedClusAvg introduces a hierarchical clustering and stratified sampling mechanism to effectively address the challenges posed by data heterogeneity.

\begin{figure}[t!]
	\centering
	\includegraphics[width=1\linewidth]{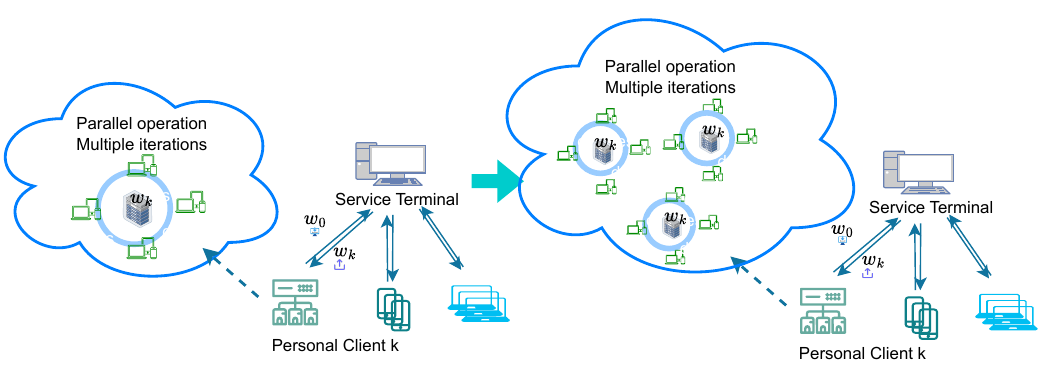}
	\caption{FedClusAvg algorithm flow}
	\label{fig:Fed}
\end{figure}

The overall workflow of the proposed FedClusAvg algorithm is illustrated in Fig.~\ref{fig:Fed}, and the corresponding pseudocode is provided in Table~\ref{tab:FCA}. Specifically, each client first performs clustering on its local dataset based on intrinsic feature attributes. This is followed by stratified sampling, which ensures that the constructed local training subset is more balanced and representative of the underlying data distribution.

At the beginning of each communication round, all clients download the latest global model parameters $\boldsymbol{w}_0$ from the central server. Each client then conducts multiple rounds of local training using the sampled data, resulting in an updated local model $\boldsymbol{w}_k$ for client $k$. Once a sufficient number of client updates $\boldsymbol{w}_k$ are collected, the server aggregates these models to produce a new global model ${\boldsymbol{w}_0}^{'}$, which is then broadcast to all clients for the next iteration.


The FedClusAvg algorithm can be used for all objective functions with regard to a finite number of sample error accumulation functions:

\begin{equation}
\min_{\boldsymbol{w} \in R^{d}} f(\boldsymbol{w})
\centering
\label{eq1}
\end{equation}

\begin{equation}
f(\boldsymbol{w})=\frac{1}{n} \sum_{i=1}^{n} f_{i}(\boldsymbol{w})
\centering
\label{eq2}
\end{equation}

\begin{equation}
f_{i}(\boldsymbol{w})=l\left(x_{i}, y_{i} ; \boldsymbol{w}_{i}\right)
\centering
\label{eq3}
\end{equation}

In federated learning, assuming that there are $K$ individual clients participating in training, $P_k$ represents the local data set stored in the $k^{th}$ client, and the sample size is $n_k$, then the objective function is:

\begin{equation}
f(\boldsymbol{w}) =\sum_{k=1}^{K} \frac{n_{k}}{n} F_{k}(\boldsymbol{w})
\centering
\label{eq4}
\end{equation}

\begin{equation}
F_{k}(\boldsymbol{w}) =\frac{1}{n_{k}} \sum_{i \in P_{k}} f_{i}(\boldsymbol{w})
\centering
\label{eq5}
\end{equation}

\begin{equation}
f_{i}(\boldsymbol{w}) =l\left(x_{i}, y_{i} ; \boldsymbol{w}_{i}\right)
\centering
\label{eq6}
\end{equation}

\textbf{Local Client-Side Operations:}

On the client side, to mitigate the Non-IID characteristics inherent in the local data distribution $P_k$, a clustering operation is performed on $P_k$. The local dataset is partitioned into multiple smaller subsets, each of which is treated as an independent sub-client. This effectively transforms the client into a local micro-server architecture. Each sub-client performs multiple rounds of local training in parallel. Subsequently, the client aggregates the learned model parameters from all sub-clients using a weighted averaging strategy to generate an updated local model, which is then transmitted to the central server for global aggregation.

To ensure effective clustering and fair comparison across features, all sample feature values are standardized to remove the influence of differing units and measurement scales. Let the local dataset on client $k$ be denoted by $x_{ki} = (x_{ki1}, x_{ki2}, \dots, x_{kip})^\mathrm{T}$, where $i = 1, 2, \dots, n_k$ and $p$ is the number of features per sample. The standardized version of each sample is computed as follows:

\begin{equation}
x_{ki}^*=\frac{x_{ki}-\overline{x_{ki}}}{\sqrt{s_{ki}}}
\centering
\label{eq7}
\end{equation}
where $x_{ki}$ is the sample mean of the sample $x_{ki}$, and $s_{ki}$ is the sample variance of the sample $x_{ki}$.

Considering that the main purpose of client-side sample clustering is to eliminate the similarity between samples and the bias of proportion, this paper uses the system clustering method based on the longest distance method to cluster the samples. The steps are as follows. And the steps are as shown  in Table \ref{tab:SC}.

\begin{itemize}
	\item Select the threshold value $\theta \left(0<\theta<1\right)$.Take any sample as the first cluster center $Z_1$, such as $Z_1=x_1$.And the sample farthest from $Z_1$ is selected as the second cluster center $Z_2$.
	
	\item Looking for a new cluster center.
		\begin{enumerate}
			\item Calculate the distance between all other samples and the existing $J$ cluster centers:$D_{i1},D_{i2}, \ldots, D_{ij} \\ \left(i=1,2, \ldots, n_k\right)$.
			
			\item  if $D_s > \theta_{D_{12}}$,then $x_{s}$ is the new cluster center, otherwise stop looking for the cluster center, where
			
			\begin{equation}
			D_{s}=\max\left\{\min_{i}\left(D_{i1}, D_{i2}, \ldots, D_{ij}\right)\right\}
			\centering
			\label{eq8}
			\end{equation}
			
			\begin{equation}
			D_{12}=\left\|Z_{1}-Z_{2}\right\|=\sqrt{\left(Z_{1}-Z_{2}\right)^{2}}
			\centering
			\label{eq9}
			\end{equation}
		\end{enumerate}
	
	\item According to the principle of nearest neighbor, all samples are assigned to the nearest cluster center:
	\begin{center}
		if $D_{il}=\min_{j} D_{ij}$, then $x_i \in Z_l$
	\end{center}
\end{itemize}

\begin{table}[htbp]
	\caption{The pseudo code of the SpectralClust algorithm}
	\label{tab:SC}
	\begin{tabular}{lcl} 
		\toprule
		\textbf{SpectralClust algorithm}  Z is the set of cluster center $P_{ki} $ is the  $k$th \\client of the  $i$th sample. $J_{k}$ is cluster set of the large client $k$ and $J_{kj}$ \\ is the sample of $J_{k}$. \\ 
		\midrule
		Give a threshold  $\theta(0<\theta<1)$ \\
		$Z \leftarrow Z_{1}\quad (Z_{1} \in P_{k}$)\\
		$Z \leftarrow Z_{2}\quad (|Z_{2}-Z_{1}|\geq|Z_{i}-Z_{1}|,Z_{i}\in P_{k} \backslash Z_{1} $) \\
		for each $Z_{j} \in Z $ do\\
		\qquad for each $P_{ki} \in P_{k} \backslash Z $ do\\
		\qquad \qquad $\boldsymbol{D}_{i j} = \left\|\boldsymbol{p}_{ki}-\boldsymbol{Z}_{j}\right\|=\sqrt{\left(\boldsymbol{p}_{ki}-\boldsymbol{Z}_{j}\right)^{2}}$\\
		for each $p_{ki} \in P_{k} \backslash Z $ do\\
		\qquad $D_{i} =\max \left\{\mathop{min}\limits_{i}\left(D_{i 1}, D_{i 2}, \ldots, D_{i j}\right)\right\}$\\
		\qquad $D_{12} = \left\|\boldsymbol{Z}_{1}-\boldsymbol{Z}_{2}\right\|=\sqrt{\left(\boldsymbol{Z}_{1}-\boldsymbol{Z}_{2}\right)^{2}}$\\
		\qquad if $D_{i} \textgreater \theta D_{12}$\\ 
		\qquad \qquad $Z \leftarrow p_{ki}$\\ 
		for each $P_{ki} \in P_{k} \backslash Z $ do\\
		\qquad  $D_{i l} == \mathop{min}\limits_{j}D_{i j}$\\
		\qquad  $ J_{kl}\leftarrow P_{ki}$\\			
		\bottomrule
	\end{tabular} 
\end{table}

To streamline the algorithmic design, we impose two simplifying assumptions on the local client dataset. First, the sample size of each client is required to exceed 100, ensuring sufficient statistical support for clustering. Second, the distance between cluster centers in the optimal classification scheme must exceed $50\%$ of the average pairwise distance between samples prior to clustering. This condition guarantees adequate separability among clusters.

The number of clusters is determined by the heuristic $\left\lfloor \frac{n}{50} \right\rfloor$, where $n$ denotes the total number of samples on the client.

Following the clustering step, each client effectively transforms into a miniature federated learning architecture. The internal algorithmic procedure executed at each client can be summarized as follows:

\begin{itemize}
	\item Small clients obtained by clustering($j$ category, the sample size of each category is $n_{kj}$)
	\begin{enumerate}
		\item Get the latest model parameter $\boldsymbol{w}_t$ from the personal client $k$; 
		\item In each small client, use the sample set $P_{kj}$ and parameter $\boldsymbol{w}_t$ in the small client to calculate the gradient $\boldsymbol{g}_{km}$ of the sample data set;
		\item Send the gradient $\boldsymbol{g}_{km}$ to the personal client $k$.
	\end{enumerate}
	
	\item Personal Client $k$
	\begin{enumerate}
		\item Obtain $\boldsymbol{g}_{k1},\boldsymbol{g}_{k2},\dots,\boldsymbol{g}_{kn_{kj}}$ from the small client, and calculate the weighted average:
		
		\begin{equation}
		\bar{\boldsymbol{g}}=\sum_{m=1}^{n_{kj}}\eta \boldsymbol{g}_{km}
		\centering
		\label{eq10}
		\end{equation}
		
		\begin{equation}
		\eta =\frac{\| \tilde{\boldsymbol{g}}-\boldsymbol{g}_{km}\|}{\sum_{m=1}^{n_{kj}} \|\tilde{\boldsymbol{g}}-\boldsymbol{g}_{km}\|}
		\centering
		\label{eq11}
		\end{equation}
		
		\begin{equation}
		\tilde{\boldsymbol{g}}=\sum_{m=1}^{n_{kj}}\frac{n_{kj}}{n_k}\boldsymbol{g}_{km}
		\centering
		\label{eq12}
		\end{equation}
		
		\item In each small client, use the sample set $P_{kj}$ and parameter $w_t$ in the small client to calculate the gradient $\boldsymbol{g}_{km}$ of the sample data set;
		\item Send the gradient $\boldsymbol{g}_{km}$ to the personal client $k$.
	\end{enumerate}
\end{itemize}

\subsubsection{Service Terminal}

On the server side, in order to minimize the impact of Non-IID data on the accuracy of the model, we propose to use the degree of parameter deviation from the average parameter as the weight of each client for the next round of model parameter update. When the degree of deviation is large, the corresponding drop reduces its weight, and when the degree of deviation is small, the weight increases.

The algorithm flow of the service terminal is roughly as follows:

\begin{enumerate}
	\item Obtain the updated parameters $\boldsymbol{w}_{(t+1)1},\boldsymbol{w}_{(t+1)2},\ldots,\\ \boldsymbol{w}_{(t+1)k}$ of each client in the current round from the personal client, and calculate the weighted average of the latest parameters value:
	
	\begin{equation}
	\boldsymbol{w}_{t+1}=\sum_{m=1}^K\eta \boldsymbol{w}_{(t+1)m}
	\centering
	\label{eq13}
	\end{equation}
	
	\begin{equation}
	\eta =\frac{\| \tilde{\boldsymbol{w}}-\boldsymbol{w}_{(t+1)m}\|}{\sum_{m=1}^{k} \|\tilde{\boldsymbol{w}}-\boldsymbol{w}_{(t+1)m}\|}
	\centering
	\label{eq14}
	\end{equation}
	
	\begin{equation}
	\quad \tilde{\boldsymbol{w}}=\sum_{m=1}^{K}\frac{n_k}{n} \boldsymbol{w}_{(t+1)m}
	\centering
	\label{eq15}
	\end{equation}

	\item Send the updated parameter $\boldsymbol{w}_{t+1}$ to each personal client for the next round of update.
\end{enumerate}

\begin{table}[h]
	\caption{The pseudo code of the FedClusAvg algorithm}
	\label{tab:FCA}
	\begin{tabular}{lcl} 
		\toprule
		\textbf{FedClusAvg algorithm}  The $K$ clients are indexed by $k$; $\alpha$ is the learning \\ rate; $ E$ is the number of iterations in each cluster set for each client;\\$ P_k$ is local dataset of the $k^{th}$ client used to update parameters; $ B$ is local\\ small batch size for client updates; $J_k$ is cluster set of the large client $k$.\\ 
		\midrule
		\textbf{Server executes: }\\
		\qquad initialize $w_0$\\
		\qquad for each round $t =1,2,\dots$ do\\
		\qquad \qquad for each client $k \in {1,2,\dots,K}$ in parallel do\\
		\vspace{5pt}
		\qquad \qquad \qquad $w_{t+1}^k \leftarrow$ ClientUpdate($k,w_t$)\\
		\vspace{5pt}
		\qquad \qquad $\tilde{w}=\sum_{m=1}^{K} \frac{n_k}{n} w_{t+1}^m$\\
		\vspace{5pt}
		\qquad \qquad $\eta^k \leftarrow \frac{\| \tilde{w}-w_{t+1}^k\|}{\sum_{k=1}^{K} \|\tilde{w}-w_{t+1}^k\|}$\\
		\vspace{10pt}
		\qquad \qquad $w_{t+1} \leftarrow \sum_{k=1}^K\eta^k w_{t+1}^k$\\
		
		\textbf{ClientUpdate($k,w$): }//Run on client $k$\\
		\qquad if $P_k>300$ do\\
		\qquad \qquad $J_k \leftarrow$ SpectralClust($P_k,J$)\\
		\qquad \qquad if $max(avg(dist(j\in J_k))<1.2avg(dist(p\in P_k))$\\
		\qquad \qquad \qquad $J_k \leftarrow J_k$\\
		\qquad \qquad \qquad else $J_k \leftarrow P_k$\\
		\qquad $B \leftarrow$ (split $J_k$ into batches of size $B$)\\
		\qquad for each class $j\in J_k $ do\\
		\qquad \qquad for each local epoch $i$ from 1 to E do\\
		\qquad \qquad \qquad for batch $b\in B$ do\\
		\qquad \qquad \qquad \qquad $w_j \leftarrow w_j-\alpha \nabla \ell(w_j,b)$\\
		\vspace{5pt}
		\qquad \qquad return $w_j$ to class $j$\\
		\vspace{5pt}
		\qquad $\bar{w} \leftarrow \sum_{j=1}^{J} \frac{n_j}{n_k} w_j$\\
		\vspace{5pt}
		\qquad $\eta_j \leftarrow \frac{\| \bar{w}-w_j\|}{\sum_{j=1}^{J} \|\bar{w}-w_j\|}$\\
		\vspace{5pt}
		\qquad $w_{t+1}^k \leftarrow \sum_{j=1}^J\eta^j w_j$\\
		\bottomrule
	\end{tabular} 
\end{table}

\subsection{FedClusAvg+ Algorithm}

In practical deployments of federated learning, servers are often distributed across multiple geographic regions. In the original FedClusAvg algorithm, each client communicates directly with a central server in a one-to-one manner, where model updates (e.g., gradients or parameters) are transmitted individually and aggregated centrally. While effective in small-scale settings, this architecture becomes inefficient as the number of clients grows.

To address scalability and communication efficiency, the enhanced FedClusAvg+ algorithm introduces a hierarchical three-tier architecture comprising clients, sub-servers, and a central server. In this design, multiple clients first communicate with a second-tier sub-server, which performs intermediate aggregation of local model updates. These aggregated results are subsequently transmitted to the central server for global averaging. Once the central model is updated, the new parameters are disseminated back to each sub-server, which then distributes them to the respective clients.

\begin{figure}[t!]
	\centering
	\includegraphics[width=1\linewidth]{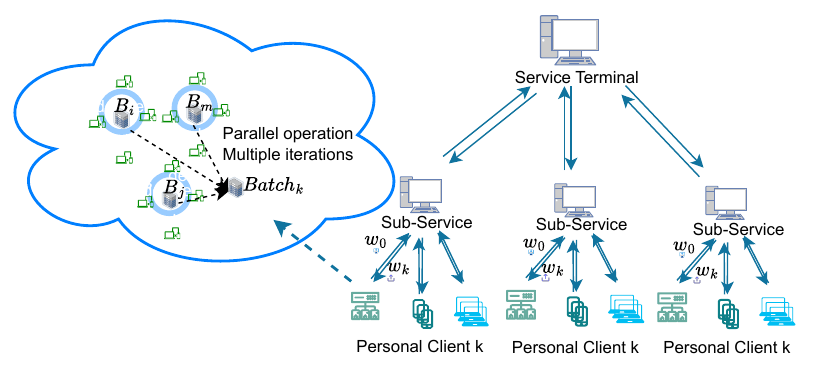}
	\caption{FedClusAvg+ algorithm}
	\label{fig:FCA1}
\end{figure}

The overall workflow of FedClusAvg+ is illustrated in Fig.~\ref{fig:FCA1}, and the corresponding pseudocode is presented in Table~\ref{tab:FCA1}.

This hierarchical communication strategy significantly reduces the number of communication rounds and alleviates bandwidth constraints, thereby enhancing overall training efficiency. In addition, by decentralizing aggregation and computation, the framework improves system robustness, mitigates the risk of single points of failure, and enhances real-time responsiveness, which is essential for False Data Injection Attack (FDIA) detection in smart grid environments.

\begin{table}[h]
	\caption{The pseudo code of the FedClusAvg+ algorithm}
	\label{tab:FCA1}
	\begin{tabular}{lcl} 
		\toprule
		\textbf{FedClusAvg+ algorithm}  The Q subservers are indexed by $q$; The $K$ clients \\of the $q^{th}$ subserver are indexed by $K_q$; $\alpha$ is the learning rate; $E$ is the \\ number of iterations in each cluster set for each client; $P_{kq}$ is local dataset \\ of the $k_q^{th}$ client used to update parameters; $B$ is local small batch size for \\ client updates; $J_kq$ is cluster set of the large client $k_q$.\\ 
		\midrule
		\textbf{Terminal Server executes: }\\
		\qquad initialize $w_0$\\
		\qquad for each round $t =1,2,\dots$ do\\
		\qquad \qquad for each subserver $q \in {1,2,\dots,Q}$ in parallel do\\
		\vspace{5pt}
		\qquad \qquad \qquad $w_{t+1}^q \leftarrow$ SubserverUpdate($Q,w_t$)\\
		\vspace{5pt}
		\qquad \qquad $\tilde{w}=\sum_{q=1}^{Q} \frac{n_q}{n} w_{t+1}^q$\\
		\vspace{5pt}
		\qquad \qquad $\eta^q \leftarrow \frac{\| \tilde{w}-w_{t+1}^q\|}{\sum_{q=1}^{Q} \|\tilde{w}-w_{t+1}^q\|}$\\
		\vspace{10pt}
		\qquad \qquad $w_{t+1} \leftarrow \sum_{q=1}^Q\eta^q w_{t+1}^q$\\
		
		\textbf{SubserverUpdate($Q,w_t$): }//Run on subserver $q$\\
		\qquad for each client $k \in {1,2,\dots,K_q}$ in parallel do\\
		\vspace{5pt}
		\qquad \qquad $w_{t+1}^k \leftarrow$ ClientUpdate($k_q,w_t$)\\
		\vspace{5pt}
		\qquad \qquad $\tilde{w^q}=\sum_{k=1}^{K_q} \frac{n_k}{n} w_{t+1}^k$\\
		\vspace{5pt}
		\qquad \qquad $\eta^{kq} \leftarrow \frac{\| \tilde{w^q}-w_{t+1}^k\|}{\sum_{k=1}^{K_q} \|\tilde{w^q}-w_{t+1}^k\|}$\\
		\vspace{10pt}
		\qquad \qquad $w_{t+1}^q \leftarrow \sum_{k=1}^{K_q}\eta^{kq} w_{t+1}^k$\\
		
		\textbf{ClientUpdate($k_q,w$): }//Run on client $k_q$\\
		\qquad if $P_{kq}>300$ do\\
		\qquad \qquad $J_{kq} \leftarrow$ SpectralClust($P_{kq},J$)\\
		\qquad \qquad if $max(avg(dist(j\in J_{kq}))<1.2avg(dist(p\in P_{kq}))$\\
		\qquad \qquad \qquad $J_{kq} \leftarrow J_{kq}$\\
		\qquad \qquad \qquad else $J_{kq} \leftarrow P_{kq}$\\
		\qquad $B \leftarrow$ (split $J_{kq}$ into batches of size $B$)\\
		\qquad for each class $j\in J_{kq} $ do\\
		\qquad \qquad for each local epoch $i$ from 1 to E do\\
		\qquad \qquad \qquad for batch $b\in B$ do\\
		\qquad \qquad \qquad \qquad $w_j \leftarrow w_j-\alpha \nabla \ell(w_j,b)$\\
		\vspace{5pt}
		\qquad \qquad return $w_j$ to class $j$\\
		\vspace{5pt}
		\qquad $\bar{w} \leftarrow \sum_{j=1}^{J} \frac{n_j}{n_k} w_j$\\
		\vspace{5pt}
		\qquad $\eta_j \leftarrow \frac{\| \bar{w}-w_j\|}{\sum_{j=1}^{J} \|\bar{w}-w_j\|}$\\
		\vspace{5pt}
		\qquad $w_{t+1}^k \leftarrow \sum_{j=1}^J\eta^j w_j$\\
		\bottomrule
	\end{tabular} 
\end{table}

\section{Experiments and Performance Evaluation} \label{l4}

To comprehensively evaluate the detection performance of the proposed FedClusAvg and FedClusAvg+ algorithms, this paper adopts seven widely-used statistical metrics: Accuracy, Precision, Recall, F-Measure (F1-score), Receiver Operating Characteristic (ROC), Area Under the Curve (AUC), and the Kolmogorov–Smirnov (KS) statistic.

Accuracy measures the proportion of correctly classified samples among all evaluated samples.  
Precision indicates the proportion of true positive samples among those predicted as positive.  
Recall (also known as sensitivity) measures the proportion of actual positive samples that are correctly identified by the model.  
F-Measure (F1-score) is the harmonic mean of Precision and Recall, providing a balanced metric when there is an uneven class distribution.  
ROC refers to the curve that plots the true positive rate (TPR) against the false positive rate (FPR) across different thresholds.  
AUC denotes the area under the ROC curve, which represents the probability that the classifier ranks a randomly chosen positive instance higher than a randomly chosen negative one.  
KS statistic measures the maximum difference between the cumulative distribution functions (CDFs) of the positive and negative samples, reflecting the model’s discriminatory power.

Suppose the confusion matrices of two classification models are shown in Table~\ref{tab:confusion}. The index is computed according to the following formula:
\begin{center}
\begin{table}[]
    \centering
    \caption{Confusion Matrix}
    \begin{tabular}{lcl} 
	\toprule
	&\textbf{Predicted positive}&\textbf{Predict negative} \\
		
	\midrule
	\textbf{Actually positive}&TP&FN\\
	\textbf{Actually negative}&FP&TN\\
	
	\bottomrule
\end{tabular} 
    \label{tab:confusion}
\end{table}
\end{center}

\addtocounter{equation}{15}
\begin{equation}
\text{Accuracy}:P_a=\dfrac{T P+T N}{T P+F P+T N+F N}
\centering
\label{eq16}
\end{equation}

\begin{equation}
\text{Precison}:P_p=\dfrac{T P}{T P+F P}
\centering
\label{eq17}
\end{equation}

\begin{equation}
\text{Recall}:R=\dfrac{T P}{T P+F N}
\centering
\label{eq18}
\end{equation}

\begin{equation}
\text{F-Measure}:\dfrac{2}{F_{1}}=\dfrac{1}{P}+\dfrac{1}{R}
\centering
\label{eq19}
\end{equation}

\begin{equation}
\text{F-Measure}:F_{1}=\dfrac{2 T P}{2 T P+F P+F N}
\centering
\label{eq20}
\end{equation}

Before describing the Area Under roc Curve, we first give two definitions, namely TPR and FPR. TPR refers to Recall, while FPR refers to the proportion of negative samples that are incorrectly predicted to be positive. The two-class model returns a probability value, and multiple points about (FPR, TPR) can be obtained by adjusting the threshold, and the resulting curve is ROC. KS is Max (TPR-FPR). The larger the KS, the better the model can separate the interval of positive and negative samples, and the better the effect of the model.

\subsection{Experimental Setup}

\textbf{Data Description:}

To verify the feasibility and effectiveness of the proposed FedClusAvg and FedClusAvg+ algorithms, we utilize the IEEE 118-Bus dataset~\cite{zimmerman2015matpower} as introduced in Section~\ref{Intro1}. In distributed energy systems, data typically originates from diverse physical devices, geographical regions, and operational states, which naturally exhibits Non-IID (Non-Independent and Identically Distributed) characteristics. The Non-IID problem is primarily manifested as an imbalance in data distribution across different buses, generators, and load nodes, leading to degraded model generalization in federated learning or data-driven analytics.

The IEEE 118-Bus dataset contains 13 input features and 1 target label. These features encompass various attributes from bus/node data, generator data, load data, branch data, and transformer data.

\textbf{Simulation Configuration:}

Following Section~\ref{Intro1}, a federated learning-based FDIA detection model is constructed based on the FedClusAvg algorithm. The model determines whether specific state variables have been maliciously compromised. After multiple rounds of feature engineering and validation, 13 informative features are selected and input into the detection model. The output indicates whether a given bus state variable has been subjected to a False Data Injection Attack (FDIA).

In the experimental setup, the grid dataset collected from various nodes is randomly partitioned into training and testing sets, comprising 60{,}000 and 10{,}000 samples, respectively. The ratio of attack to normal samples is maintained at 2:8 to simulate realistic intrusion scenarios. To reflect the non-independent and identically distributed (Non-IID) nature of data in federated environments, the training set is divided into 100 client datasets using a label skewing strategy, where the proportion of attack samples varies across clients.

FDIA detection is performed using the Rec-AD model we proposed in reference~\cite{li2025rec}, integrated within the FedClusAvg framework. The GPU configuration is the same as Rec-AD in our experiment.

All model weights are initialized to zero, and the learning rate is set to 0.01. Both the FedClusAvg and FedAvg algorithms are implemented using the FedML framework. Experiments are executed in the Visual Studio Code (VSCode) environment with Python extensions, running on Ubuntu 20.04.4 LTS and Python 3.8. The software dependencies include PyTorch 1.13 and FedML 0.7.1. A summary of the experimental configuration is provided in Table~\ref{tab:test}.

\begin{table}[htb]
\centering
	\caption{\label{tab:test}Parameter settings}
 \scalebox{0.9}{
 \begin{tabular}{ccl}
		\toprule
		\bf{Parameter} & \bf{Parameter Value}	& \bf{Parameter Annotations} \\
		\midrule
		k & 100 & Number of clients by dataset\\
		
		\cmidrule(lr){1-3}
		Q & 5 &	Number of sub-servers\\
		
		\cmidrule(lr){1-3}
		$\alpha$ & 0.01 &	Learning Rate\\
		\cmidrule(lr){1-3}
		E & 10 & The number of times each client trains \\
        &&its local dataset in each round\\
		
		\cmidrule(lr){1-3}
		M & 400 & Total number of updates on the server\\
		\bottomrule
	\end{tabular}
 }
\end{table}

\subsection{Performance Analysis}

In the first part of the experiment, we compare the performance of the FedClusAvg and FedAvg algorithms under identical federated learning conditions. Both models are trained for 400 iterations using the same training dataset. The resulting parameters are then evaluated on the test set to assess model effectiveness. A comparative analysis of the two algorithms is presented in Table~\ref{tab:Indicator}, which reports the descriptive statistics for four key evaluation metrics: Accuracy, Precision, Recall, and F1-Score. For each metric, five statistical values are provided: minimum, first quartile (Q1), median, mean, and maximum.

\begin{figure}[t!]
	\centering
	\includegraphics[width=1\linewidth]{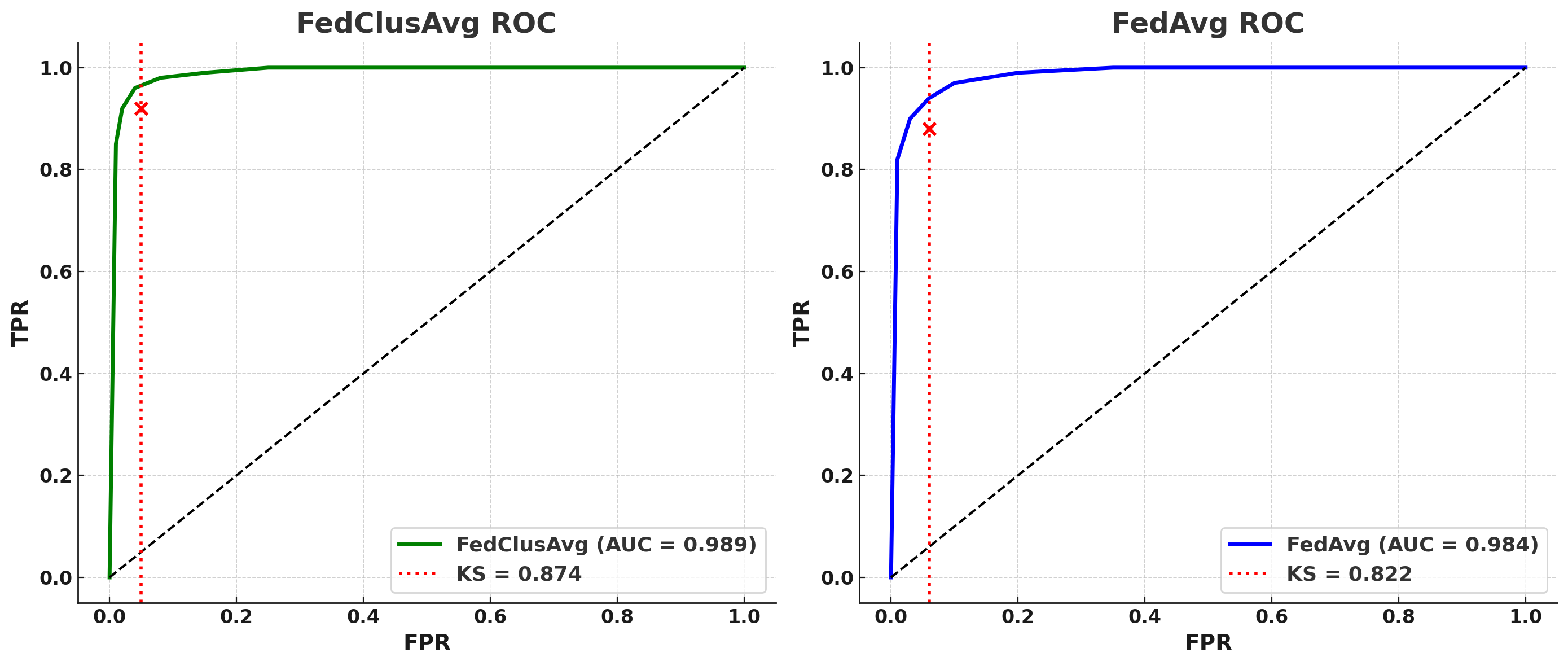}
	\caption{Model KS values by FedClusAvg (left) algorithm and FedAvg (right) algorithm }
\label{fig:token_cluster}
\end{figure}

The Receiver Operating Characteristic (ROC) curves corresponding to the two algorithms are depicted in Fig~\ref{fig:token_cluster}. As shown, the Kolmogorov–Smirnov (KS) statistic of the FedClusAvg algorithm exceeds that of FedAvg, indicating superior discriminative performance. The KS statistic measures the maximum difference between the true positive rate (TPR) and the false positive rate (FPR), with higher values indicating better separation between positive and negative classes.

\begin{table}[t!]\small	
\centering
	\caption{\label{tab:test1} Evaluation metrics of FedClusAvg and FedAvg}   
\scalebox{0.9}{ \begin{tabular}{lrrrrrr}	
		\toprule
		& \bf{Min.} & \bf{1st Qu} & \bf{Med} & \bf{Mean}  &\bf{Max}\\
		\midrule
		FedClusAvg&\multirow{2}{*}{0.9423}&\multirow{2}{*}{0.9512}&\multirow{2}{*}{0.9568}&\multirow{2}{*}{0.9571}&\multirow{2}{*}{0.9687}\\
		Accuracy&~&~&~&~&~&~\\
		
		\cmidrule(lr){1-7}
		FedAvg&\multirow{2}{*}{0.9187}&\multirow{2}{*}{0.9254}&\multirow{2}{*}{0.9321}&\multirow{2}{*}{0.9318}&\multirow{2}{*}{0.9427}\\
		Accuracy&~&~&~&~&~&~\\
		
		\cmidrule(lr){1-7}
		FedClusAvg&\multirow{2}{*}{0.9382}&\multirow{2}{*}{0.9475}&\multirow{2}{*}{0.9539}&\multirow{2}{*}{0.9542}&\multirow{2}{*}{0.9663}\\
		Precison&~&~&~&~&~&~\\
		\cmidrule(lr){1-7}
		FedAvg&\multirow{2}{*}{0.9135}&\multirow{2}{*}{0.9208}&\multirow{2}{*}{0.9267}&\multirow{2}{*}{0.9271}&\multirow{2}{*}{0.9385}\\
		Precison&~&~&~&~&~&~\\
		\cmidrule(lr){1-7}
		FedClusAvg&\multirow{2}{*}{0.9401}&\multirow{2}{*}{0.9487}&\multirow{2}{*}{0.9552}&\multirow{2}{*}{0.9558}&\multirow{2}{*}{0.9671}\\
		Recall&~&~&~&~&~&~\\
		\cmidrule(lr){1-7}
		FedAvg&\multirow{2}{*}{0.9162}&\multirow{2}{*}{0.9231}&\multirow{2}{*}{0.9289}&\multirow{2}{*}{0.9294}&\multirow{2}{*}{0.9406}\\
		Recall&~&~&~&~&~&~\\
		\cmidrule(lr){1-7}
		FedClusAvg&\multirow{2}{*}{0.9391}&\multirow{2}{*}{0.9481}&\multirow{2}{*}{0.9545}&\multirow{2}{*}{0.9550}&\multirow{2}{*}{0.9667}\\
		$F_{1}$&~&~&~&~&~&~\\
		\cmidrule(lr){1-7}
		FedAvg&\multirow{2}{*}{0.9148}&\multirow{2}{*}{0.9219}&\multirow{2}{*}{0.9278}&\multirow{2}{*}{0.9282}&\multirow{2}{*}{0.9395}\\
		$F_{1}$&~&~&~&~&~&~\\
		\bottomrule
	\end{tabular}} 	
 \label{tab:Indicator}
\end{table}

\begin{figure}[t!]
	\centering
	\includegraphics[width=1\linewidth]{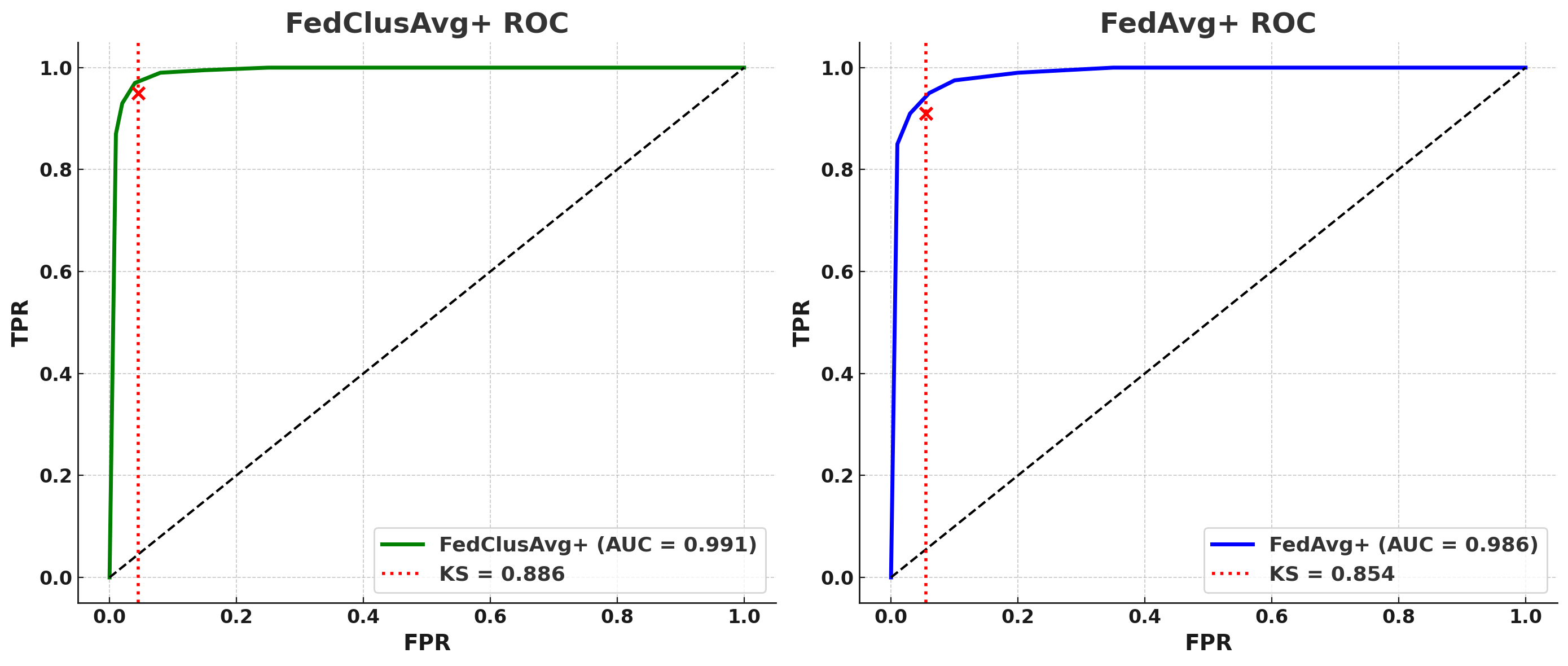}
	\caption{Model KS values obtained by FedClusAvg+ (left) and FedAvg+ (right) algorithms}
	\label{fig:token_cluster1}
\end{figure}

In the second part of the experiment, the FedClusAvg+ and FedAvg+ algorithms are evaluated under the same settings as in the previous section, with the only modification being the introduction of the three-tier hierarchical communication parameter $Q$. The descriptive statistics for the evaluation metrics are summarized in Table~\ref{tab:indicator1}, while the corresponding ROC curves are illustrated in Fig~\ref{fig:token_cluster1}. As shown, the Kolmogorov–Smirnov (KS) statistic of the FedClusAvg+ algorithm exceeds that of FedAvg+, indicating stronger classification capability.

A notable observation is that the Recall values of both FedAvg and FedAvg+ are consistently lower than those achieved by FedClusAvg and FedClusAvg+. This discrepancy is likely attributable to the pronounced effects of data heterogeneity. Since FedAvg and FedAvg+ apply uniform averaging across all clients without accounting for the underlying Non-IID characteristics, their global models often fail to generalize well across diverse data distributions. Specifically, when attack patterns are unevenly distributed across clients, the global model tends to underperform in detecting minority attack cases, resulting in lower recall.

Moreover, the absence of an advanced aggregation strategy in FedAvg+ may lead to ineffective integration of client-specific knowledge. In contrast, both FedClusAvg and FedClusAvg+ leverage a clustering-based parameter aggregation mechanism that better captures structural variations in local data. This enables the global model to generalize more effectively across heterogeneous clients, thereby achieving higher recall and overall performance in FDIA detection tasks.

\begin{table}[t!]\small	
\centering
\caption{\label{tab:test2}  Evaluation metrics of FedClusAvg+ and FedAvg+}   
\scalebox{0.9}{\begin{tabular}{lrrrrrr}	
		\toprule
		& \bf{Min.} & \bf{1st Qu.} & \bf{Med} & \bf{Mean}  &\bf{Max.}\\
		\midrule
		FedClusAvg+&\multirow{2}{*}{0.9503}&\multirow{2}{*}{0.9572}&\multirow{2}{*}{0.9638}&\multirow{2}{*}{0.9632}&\multirow{2}{*}{0.9749}\\
		Accuracy&~&~&~&~&~&~\\
		
		\cmidrule(lr){1-7}
		FedAvg+&\multirow{2}{*}{0.9346}&\multirow{2}{*}{0.9398}&\multirow{2}{*}{0.9452}&\multirow{2}{*}{0.9457}&\multirow{2}{*}{0.9568}\\
		Accuracy&~&~&~&~&~&~\\
		
		\cmidrule(lr){1-7}
		FedClusAvg+&\multirow{2}{*}{0.9467}&\multirow{2}{*}{0.9539}&\multirow{2}{*}{0.9611}&\multirow{2}{*}{0.9604}&\multirow{2}{*}{0.9730}\\
		Precison&~&~&~&~&~&~\\
		\cmidrule(lr){1-7}
		FedAvg+&\multirow{2}{*}{0.9304}&\multirow{2}{*}{0.9359}&\multirow{2}{*}{0.9418}&\multirow{2}{*}{0.9423}&\multirow{2}{*}{0.9542}\\
		Precison&~&~&~&~&~&~\\
		\cmidrule(lr){1-7}
		FedClusAvg+&\multirow{2}{*}{0.9484}&\multirow{2}{*}{0.9554}&\multirow{2}{*}{0.9625}&\multirow{2}{*}{0.9619}&\multirow{2}{*}{0.9743}\\
		Recall&~&~&~&~&~&~\\
		\cmidrule(lr){1-7}
		FedAvg+&\multirow{2}{*}{0.9323}&\multirow{2}{*}{0.9376}&\multirow{2}{*}{0.9443}&\multirow{2}{*}{0.9438}&\multirow{2}{*}{0.9553}\\
		Recall&~&~&~&~&~&~\\
		\cmidrule(lr){1-7}
		FedClusAvg+&\multirow{2}{*}{0.9475}&\multirow{2}{*}{0.9546}&\multirow{2}{*}{0.9618}&\multirow{2}{*}{0.9611}&\multirow{2}{*}{0.9736}\\
		$F_{1}$&~&~&~&~&~&~\\
		\cmidrule(lr){1-7}
		FedAvg+&\multirow{2}{*}{0.9313 }&\multirow{2}{*}{0.9367 }&\multirow{2}{*}{0.9425}&\multirow{2}{*}{0.9430 }&\multirow{2}{*}{0.9547}\\
		$F_{1}$&~&~&~&~&~&~\\
		\bottomrule
	\end{tabular}	
    }
 \label{tab:indicator1}
\end{table}	

\begin{table}[htbp]\small
\centering
\caption{Comparison of evaluation indicators of FedClusAvg+, FedProx and FedNova}
\scalebox{0.9}{\begin{tabular}{lccccc}
\toprule
 & \textbf{Min.} & \textbf{1st Qu.} & \textbf{Med} & \textbf{Mean} & \textbf{Max} \\
\midrule
FedClusAvg+ Accuracy & 0.9503 & 0.9572 & 0.9638 & 0.9632 & 0.9749 \\
FedProx Accuracy         & 0.9302 & 0.9375 & 0.9440 & 0.9445 & 0.9561 \\
FedNova Accuracy         & 0.9251 & 0.9322 & 0.9393 & 0.9396 & 0.9507 \\
\midrule
FedClusAvg+ Precision & 0.9467 & 0.9539 & 0.9611 & 0.9604 & 0.9730 \\
FedProx Precision        & 0.9320 & 0.9401 & 0.9472 & 0.9478 & 0.9584 \\
FedNova Precision        & 0.9247 & 0.9328 & 0.9396 & 0.9401 & 0.9513 \\
\midrule
FedClusAvg+ Recall    & 0.9484 & 0.9554 & 0.9625 & 0.9619 & 0.9743 \\
FedProx Recall           & 0.9342 & 0.9424 & 0.9501 & 0.9505 & 0.9632 \\
FedNova Recall           & 0.9261 & 0.9341 & 0.9418 & 0.9421 & 0.9554 \\
\midrule
FedClusAvg+ F1        & 0.9475 & 0.9546 & 0.9618 & 0.9611 & 0.9736 \\
FedProx F1               & 0.9331 & 0.9414 & 0.9484 & 0.9489 & 0.9605 \\
FedNova F1               & 0.9258 & 0.9337 & 0.9402 & 0.9408 & 0.9523 \\
\bottomrule
\end{tabular}
}
\label{tab:fedclusavg_plus_vs_prox_nova}
\end{table}

Analysis of the aforementioned evaluation metrics reveals that both FedClusAvg and FedClusAvg+ consistently outperform their baseline counterparts across all evaluation criteria. Moreover, the proposed algorithms demonstrate superior stability, thereby enhancing the real-time responsiveness and robustness of FDIA detection. By mitigating regional misclassification risks, they significantly contribute to the overall security and intelligence of modern smart grid infrastructures.

To further assess the advantages of the proposed hierarchical communication framework, we conducted comparative experiments on the IEEE 118-bus system dataset, evaluating FedClusAvg+ against FedProx~\cite{MLSYS2020_1f5fe839} and FedNova~\cite{wang2020tackling} for FDIA detection. FedProx introduces a proximal term to reduce the impact of client-side model drift, while FedNova applies normalized update aggregation to stabilize convergence. Despite their design improvements, both methods show limited performance gains under heterogeneous conditions.

The experimental results demonstrate that FedClusAvg+ outperforms FedProx and FedNova across all metrics, including Accuracy, Precision, Recall, and F1-score. In particular, FedClusAvg+ achieves an average Accuracy of 96.32\% and an F1-score of 96.11\%, compared to 94.45\% and 94.89\% for FedProx, and 93.96\% and 94.08\% for FedNova, respectively. These results highlight the effectiveness of the multi-tier architecture in addressing Non-IID data challenges and improving overall detection performance.

Furthermore, FedClusAvg+ retains the clustering-based aggregation advantages of FedClusAvg while introducing sub-server layers to intermediate client-server communication. This hierarchical architecture alleviates communication bottlenecks at the central server, enhances training throughput, and improves scalability in large-scale federated deployments. These characteristics make FedClusAvg+ particularly suitable for real-world applications in distributed and latency-sensitive environments such as smart grids.

\subsection{Generalization Analysis}
\begin{table}[htbp]\small
\centering
\caption{Test system size}
\scalebox{0.65}{
\begin{tabular}{lrrrr}
\toprule
\textbf{System size} & \textbf{Number of } & \textbf{Number of } & \textbf{Number of } & \textbf{Number of } \\
&\textbf{nodes}&\textbf{measurement points}&\textbf{training samples}&\textbf{test samples} \\
\hline
IEEE 118 & 118 & 490 & 50,000 & 10,000 \\
IEEE 300 & 300 & 1,122 & 120,000 & 25,000 \\
Regional power grid & 1,500 & 5,800 & 500,000 & 100,000 \\
Provincial power grid & 5,000 & 22,000 & 1,200,000 & 250,000 \\
\bottomrule
\end{tabular}
}
\label{table:size}
\end{table}


\begin{table}[htbp]\small
\centering
\caption{Comparison of Accuracy and AUC of different federated algorithms at different system scales}
\scalebox{0.6}{\begin{tabular}{lcccccccc}
\toprule
\multirow{2}{*}{\textbf{System size}} &
\multicolumn{2}{c}{\textbf{FedAvg+}} &
\multicolumn{2}{c}{\textbf{FedClusAvg+}} &
\multicolumn{2}{c}{\textbf{FedProx}} &
\multicolumn{2}{c}{\textbf{FedNova}} \\
\cmidrule(r){2-3} \cmidrule(r){4-5} \cmidrule(r){6-7} \cmidrule(r){8-9}
 & Accuracy & AUC & Accuracy & AUC & Accuracy & AUC & Accuracy & AUC \\\\
\midrule
IEEE118-Bus & 94.5\% & 0.946 & 96.3\% & 0.965 & 95.0\% & 0.951 & 94.3\% & 0.943 \\
IEEE300-Bus & 93.4\% & 0.933 & 95.1\% & 0.952 & 93.9\% & 0.938 & 93.5\% & 0.931 \\
Regional power grid    & 91.5\% & 0.928 & 94.3\% & 0.932 & 92.1\% & 0.933 & 91.7\% & 0.926 \\
Provincial power grid    & 90.3\% & 0.911 & 92.8\% & 0.921 & 90.8\% & 0.919 & 90.5\% & 0.913 \\
\bottomrule
\end{tabular}
}
\label{tab:fed_methods_comparison}
\end{table}

Table~\ref{table:size} summarizes the basic characteristics of the power systems used in this study, ranging from the standard IEEE 118-bus system to full-scale provincial power grids. This design enables a comprehensive evaluation of the scalability and robustness of the proposed algorithms. As system size increases, the number of nodes and measurements grows approximately linearly, and the volume of training and testing samples is scaled accordingly to ensure sufficient data diversity and representativeness.

Table~\ref{tab:fed_methods_comparison} compares the baseline detection performance of FedAvg+ and FedClusAvg+ across different system scales. The results consistently show that FedClusAvg+ outperforms FedAvg+ in all scenarios. Notably, in the provincial grid setting, FedClusAvg+ achieves an Accuracy of 92.8\%, which is 2.5 percentage points higher than FedAvg+, indicating enhanced generalization and robustness in large-scale systems. Furthermore, comparisons with FedProx and FedNova confirm that FedClusAvg+ achieves superior performance across all evaluated benchmarks.

Across the four test systems, FedClusAvg+ attains an average Accuracy of 94.63\% and AUC of 0.9425, significantly outperforming alternative methods. This advantage is attributed to its hierarchical communication framework and client-side clustering strategy, which jointly mitigate communication latency, model drift, and data heterogeneity. Among the baselines, FedProx generally surpasses FedAvg+ by introducing a proximal term that constrains the divergence between local and global models. This yields Accuracy and AUC gains of approximately 0.6\%–1.0\% and 0.005–0.008, respectively—particularly notable in heterogeneous environments like regional or provincial grids.

FedNova demonstrates performance comparable to FedAvg+, with limited improvements in detection accuracy. Although its communication normalization offers theoretical convergence benefits, it does not translate into meaningful gains in FDIA detection performance. In some scenarios, FedNova even underperforms FedAvg+, suggesting that it is better suited for correcting unbalanced local updates rather than enhancing detection capability.

As system scale increases, the performance disparities among algorithms become more pronounced. While the differences are marginal in small-scale systems (e.g., IEEE 118/300-bus), they become significant in more complex settings, where communication architecture, aggregation strategy, and robustness are critical to detection effectiveness.

Table~\ref{table:performance} presents the detection performance of FedAvg+ and FedClusAvg+ against various FDIA attack types on the IEEE 300-bus system. FedClusAvg+ consistently achieves higher Recall and lower False Alarm Rate (FAR) across all attack scenarios. Specifically, under progressive FDIA—the most stealthy attack type—FedClusAvg+ attains a Recall of 94.1\%, 3.8 percentage points higher than FedAvg+, while reducing FAR by 2.9 percentage points. These results highlight the ability of FedClusAvg+ to capture subtle attack patterns more effectively, owing to its clustering and attention mechanisms.

Table~\ref{table:unbalance} evaluates performance under class-imbalanced conditions (attack:normal = 1:9) on the IEEE 300-bus system. The advantages of FedClusAvg+ become more prominent under this imbalance, with improvements of 8.2\% in Precision, 6.3\% in Recall, and 7.3\% in F1-score over FedAvg+. These gains underscore the efficacy of hierarchical clustering and adaptive aggregation in handling minority class detection, maintaining low false positive rates, and ensuring practical applicability in real-world grid monitoring.

Table~\ref{table:fitting} investigates model adaptability under varying levels of client heterogeneity. As heterogeneity increases, both methods experience performance degradation; however, FedClusAvg+ exhibits substantially greater robustness. Under high heterogeneity (70\% variation), FedClusAvg+ maintains an accuracy of 89.7\% on the IEEE 300-bus system—8.5 percentage points higher than FedAvg+. On the provincial grid, this gap widens to 9.9 percentage points. These results validate the effectiveness of FedClusAvg+'s hierarchical clustering in addressing data distribution disparities and demonstrate its suitability for large-scale, resource-diverse power system environments.

\begin{table}[htbp]
\centering
\caption{Detection performance of different attack types on IEEE300-Bus system}
\scalebox{0.85}{
\begin{tabular}{lrrrr}
\toprule
\textbf{Attack Types} & \textbf{FedAvg+} & \textbf{FedClusAvg+} & \textbf{FedAvg+} & \textbf{FedClusAvg+} \\
 & \textbf{Recall} & \textbf{Recall} & \textbf{FAR} & \textbf{FAR} \\
\hline
Random FDIA & 96.2\% & 97.5\% & 5.3\% & 3.8\% \\
Collaborative FDIA & 93.8\% & 95.2\% & 6.5\% & 4.2\% \\
Hidden FDIA & 91.5\% & 94.8\% & 7.8\% & 5.1\% \\
Intermittent FDIA & 92.7\% & 95.6\% & 6.9\% & 4.5\% \\
Progressive FDIA & 90.3\% & 94.1\% & 8.2\% & 5.3\% \\
\bottomrule
\end{tabular}
}
\label{table:performance}
\end{table}

\begin{table}[htbp]\small
\centering
\caption{Unbalanced data representation on IEEE300-Bus system (attack:normal=1:9)}
\scalebox{0.85}{
\begin{tabular}{lrrr}
\toprule
\textbf{Index} & \textbf{FedAvg+} & \textbf{FedClusAvg+} & \textbf{Promote} \\
\hline
Accuracy & 92.3\% & 95.1\% & +2.8\% \\
Precision & 78.5\% & 86.7\% & +8.2\% \\
Recall & 83.2\% & 89.5\% & +6.3\% \\
F1 & 80.8\% & 88.1\% & +7.3\% \\
AUC & 0.912 & 0.945 & +0.033 \\
\bottomrule
\end{tabular}
}
\label{table:unbalance}
\end{table}

\begin{table}[htbp]\small
\centering
\caption{Heterogeneous client adaptability}
\scalebox{0.85}{
\begin{tabular}{lrrrr} 
\toprule
\textbf{Heterogeneity} & \textbf{FedAvg+} & \textbf{FedClusAvg+} & \textbf{FedAvg+} & \textbf{FedClusAvg+} \\
 & \textbf{300-Bus} & \textbf{300-Bus} & \textbf{Provincial} & \textbf{Provincial} \\
\hline
Homogeneous & 93.8\% & 95.7\% & 90.3\% & 93.8\% \\
&&&&\\
Low het. & 93.2\% & 95.1\% & 88.5\% & 93.2\% \\
(10\% difference)&&&&\\
Medium het. & 90.5\% & 94.8\% & 85.2\% & 91.5\% \\
 (30\% difference)&&&&\\
High het. & 86.3\% & 92.5\% & 80.7\% & 88.9\% \\
(50\% difference)&&&&\\
Very high het. & 81.2\% & 89.7\% & 75.3\% & 85.2\% \\
(70\% difference)&&&&\\
\bottomrule
\end{tabular}
}
\label{table:fitting}
\end{table}

\subsection{Communication efficiency analysis}
\begin{figure}[htb]
	\centering
	\includegraphics[width=1.0\linewidth]{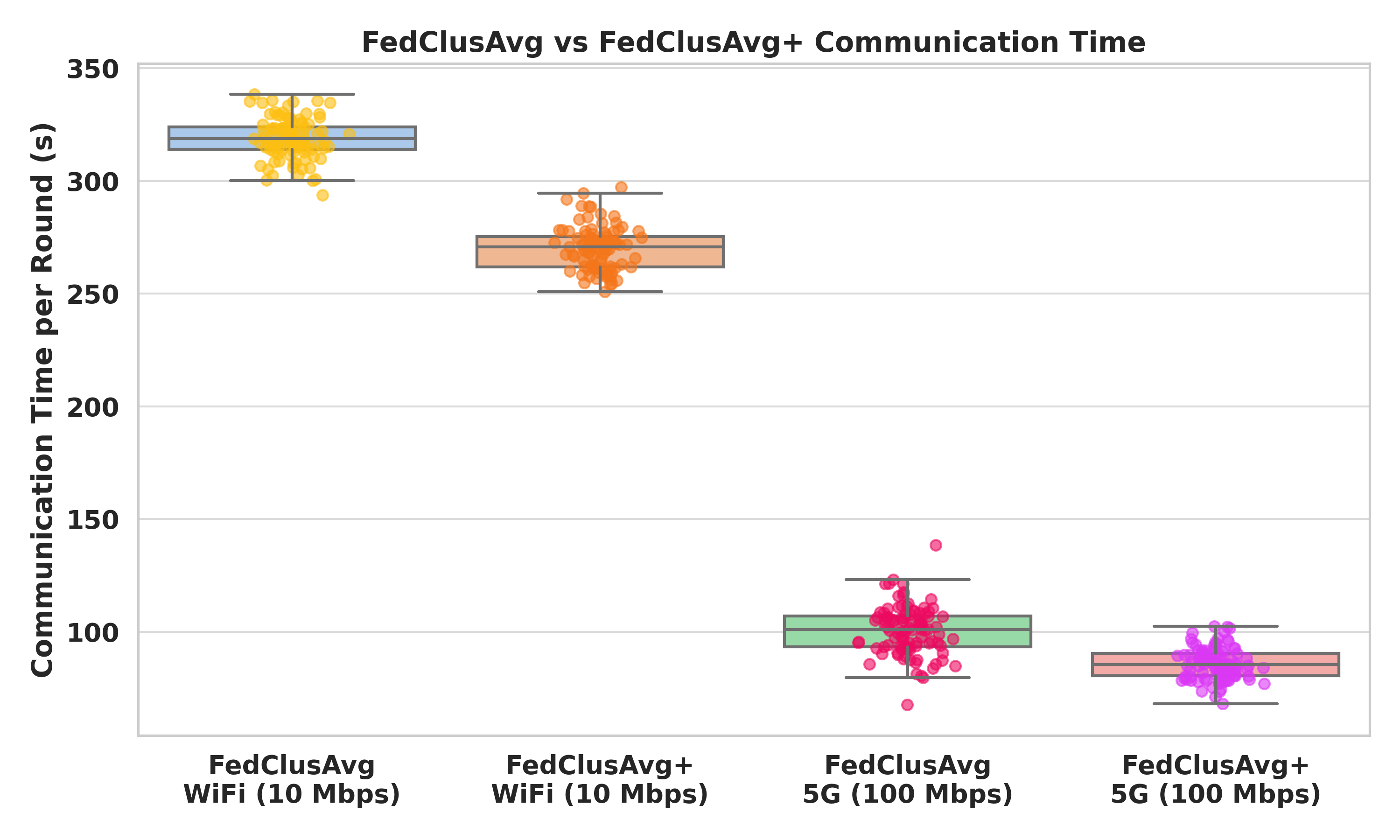}
	\caption{FedClusAvg vs FedClusAvg+ Communication efficiency }
	\label{fig:Output}
\end{figure}

In the FDIA detection task based on the IEEE 118-bus dataset, we conduct a comparative analysis of the communication efficiency and system scalability between the conventional Federated Clustered Averaging algorithm (FedClusAvg) and its enhanced variant with a three-tier communication architecture (FedClusAvg+).

In terms of communication efficiency, FedClusAvg+ exhibits notable advantages over FedClusAvg. The original FedClusAvg employs a standard client–server structure, wherein each communication round involves all clients uploading their local model updates to a central server and subsequently downloading the updated global model. As the number of clients increases or network bandwidth becomes limited, this centralized communication pattern can lead to severe congestion, especially under low-bandwidth environments such as WiFi.

To address this limitation, FedClusAvg+ introduces an intermediate layer of sub-servers, forming a hierarchical client–sub-server–central server communication topology. During each communication round, clients first transmit their locally trained models to designated sub-servers. These sub-servers perform local aggregation and forward the summarized updates to the central server. By reducing direct client-to-central server communication, this architecture significantly alleviates communication bottlenecks and enables parallel processing across multiple sub-servers, thereby decreasing overall system latency.

As illustrated in Fig.~\ref{fig:Output}, empirical results on the IEEE 118-bus dataset demonstrate that under WiFi conditions (10 Mbps), FedClusAvg incurs an average communication delay of approximately 320 seconds per round, while FedClusAvg+ reduces this delay to around 270 seconds—yielding a reduction of roughly 15.6\%. Under 5G conditions (100 Mbps), the advantage becomes more pronounced: FedClusAvg+ achieves a per-round delay of 85 seconds, compared to 100 seconds for FedClusAvg. These findings highlight the superior bandwidth utilization and network adaptability of the hierarchical FedClusAvg+ design. Moreover, FedClusAvg+ supports enhanced concurrency by enabling multiple sub-servers to concurrently aggregate updates from different subsets of clients. This mitigates the congestion typically observed when numerous clients simultaneously communicate with a single central server. Consequently, FedClusAvg+ not only improves communication efficiency and reduces latency but also provides a more scalable foundation for deployment in large-scale federated environments.

In summary, the hierarchical communication design of FedClusAvg+ offers clear improvements in both communication efficiency and system scalability, rendering it particularly well-suited for collaborative FDIA detection in large-scale distributed power systems.

\section{Conclusion} \label{l6}

We proposed an enhanced federated learning framework based on the FedClusAvg algorithm to address key challenges in False Data Injection Attack (FDIA) detection for smart grids. The framework tackles the Non-IID nature of distributed measurements and preserves data privacy through a hierarchical architecture that integrates clients, sub-servers, and a central server. By incorporating clustered sampling, deviation-based weighted aggregation, and multi-tier communication, the proposed method improves model generalization, detection accuracy, and training efficiency. It significantly reduces communication rounds and bandwidth consumption while enhancing robustness against regional misclassification and adversarial exploitation of data heterogeneity.
Experimental evaluations on the IEEE 118-bus and 300-bus systems demonstrate that the proposed algorithm consistently outperforms classical methods such as FedAvg, FedProx, and FedNova in terms of accuracy, recall, and communication efficiency. Moreover, the framework ensures privacy compliance and reduces the risk of data breaches, making it suitable for real-world deployment in large-scale, heterogeneous, and security-critical power systems.
In summary, the proposed approach advances the development of secure, scalable, and privacy-preserving FDIA detection solutions in modern smart grids.

\bibliography{Test-bibtex}

\bibliographystyle{IEEEtran}

\end{document}